\documentclass{article}

\usepackage[final]{corl_2023} 

\usepackage{multirow}
\usepackage{bbding}
\usepackage{pifont}
\usepackage[nointegrals]{wasysym}
\usepackage{caption}
\usepackage{subcaption}
\usepackage{array}
\usepackage{siunitx}
\usepackage{color}
\usepackage[normalem]{ulem}
\usepackage{hyperref}
\usepackage{nomencl}
\makenomenclature
\usepackage{algorithm, setspace}
\usepackage{algpseudocode}%
\usepackage{amsmath,amssymb,amsfonts}
\usepackage{graphicx}
\usepackage{textcomp}
\usepackage{wrapfig}
\usepackage{mdframed,lipsum}
\usepackage[T1]{fontenc} 
\usepackage{booktabs}
\usepackage{mwe}
\usepackage{array,booktabs}
\usepackage[table]{xcolor}

\usepackage{appendix}
\usepackage{etoolbox}

\newcommand*{\affaddr}[1]{#1} 
\newcommand*{\affmark}[1][*]{\textsuperscript{#1}}
\newcommand*{\email}[1]{\texttt{#1}}

\title{Multimodal Learning of Soft Robot Dynamics using Differentiable Filters}

\author{%
Xiao Liu\affmark[1], Yifan Zhou\affmark[1], Shuhei Ikemoto\affmark[2], and Heni Ben Amor\affmark[1]\\
\affaddr{\affmark[1]Interactive Robotics Lab, Arizona State University}\\
\affaddr{\affmark[2]Kyushu Institute of Technology}\\
\affmark[1]\email{\{xliu330,yzhou298,hbenamor\}@asu.edu}\quad \affmark[2]\email{ikemoto@brain.kyutech.ac.jp}\\
}

\begin{document}
\maketitle

\begin{abstract}

Differentiable Filters, as recursive Bayesian estimators, possess the ability to learn complex dynamics by deriving state transition and measurement models exclusively from data. This data-driven approach eliminates the reliance on explicit analytical models while maintaining the essential algorithmic components of the filtering process. However, the gain mechanism remains non-differentiable, limiting its adaptability to specific task requirements and contextual variations.
To address this limitation, this paper introduces an innovative approach called $\alpha$-MDF ({\bf A}ttention-based {\bf M}ultimodal {\bf D}ifferentiable {\bf F}ilter). $\alpha$-MDF leverages modern attention mechanisms to learn multimodal latent representations for accurate state estimation in soft robots. By incorporating attention mechanisms, $\alpha$-MDF offers the flexibility to tailor the gain mechanism to the unique nature of the task and context.
The effectiveness of $\alpha$-MDF is validated through real-world state estimation tasks on soft robots. Our experimental results demonstrate significant reductions in state estimation errors, consistently surpassing differentiable filter baselines by up to 45\% in the domain of soft robotics.

\end{abstract}

\keywords{Multimodal Learning, Sensor Fusion, Differentiable Filters.  } 


\section{Introduction}

\begin{wrapfigure}{r}{0.5\textwidth}
\vspace{-0.1in}
  \begin{center}
    \includegraphics[width=0.5\textwidth]{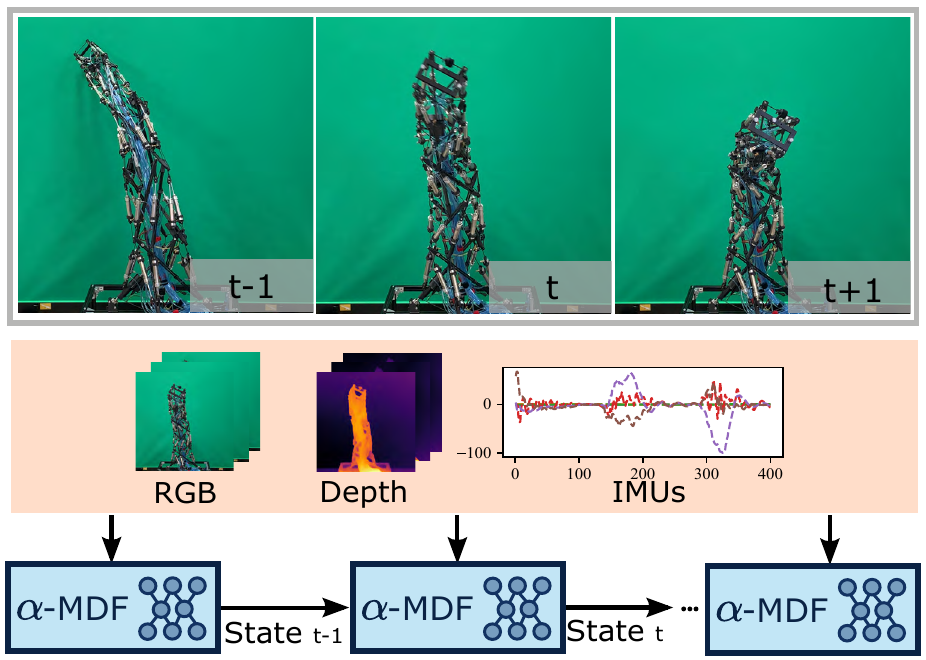}
  \end{center}
  \caption{The $\alpha$-MDF framework enables state estimation in multimodal settings applicable to soft robots.}
  \label{fig:teaser}
  \vspace{-0.2in}
\end{wrapfigure}
Soft robots, characterized as deformable structures capable of actuation, are composed of materials that facilitate smooth curved shapes~\cite{lee2017soft}. They possess remarkable capabilities in executing a wide spectrum of movements~\cite{yumbla2021human}, including extension, contraction, bending, shearing, and twisting~\cite{lee2020twister}, rendering them remarkably adaptable for maneuvering through confined spaces encountered in medical~\cite{burgner2015continuum} and industrial contexts. Notably, the integration of tensegrity structures~\cite{skeleton2009tensegrity}, featuring compressive members supported by tensile cables, has emerged as a prominent design approach. This integration effectively combines the inherent flexibility of soft systems with the advantageous properties of rigid components~\cite{jung2018bio, kim2020rolling, ikemoto2021development}. Tensegrity robots exhibit the ability to resist compressive forces in specific directions while maintaining overall flexibility.
Nevertheless, the complexity of learning soft robot dynamics presents formidable challenges in state estimation. The highly nonlinear nature and extensive degrees of freedom inherent in these systems further compound the difficulties associated with accurate modeling and prediction.
A novel perspective for learning soft robot dynamics is to derive the underlying models from data alone. Recent advancements in Deep state-space models (DSSMs)~\cite{NEURIPS2018_5cf68969} provide effective solutions for understanding the state and measurement estimation from observed sequences as data-driven approaches~\cite{NEURIPS2018_5cf68969, klushyn2021latent, kloss2021train, liu2023enhancing, liu2023learning,weigend2023probabilistic}. Such approaches do not need to derive explicit system dynamics, which is essential and challenging in traditional filtering techniques. A subclass of algorithms derived from DSSMs, called Differentiable Filters (DFs), focus on learning state transition and measurement models from data while retaining the fundamental principles of Bayesian recursive filtering as shown in Fig.~\ref{fig:teaser}. This combination of properties renders DFs particularly well-suited for soft robot systems with complex dynamics and diverse sensor observations.

\begin{figure*}[t!]
\centering
\includegraphics[width=\linewidth]{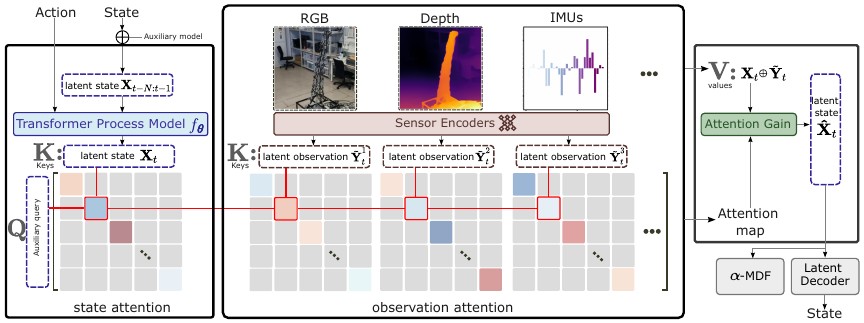}
\caption{The $\alpha$-MDF framework consists of a transformer process model, sensor encoders, and an attention gain update model. The transformer process model predicts the following latent state, while the sensor encoders learn latent representations from observations. The attention gain model then corrects the predicted latent state using the learned representations.}
\vspace{-1\baselineskip}
\label{fig:overview}
\end{figure*}

In this paper, we present a novel class of differentiable filters utilizing neural attention mechanisms. Our key innovation lies in replacing the traditional Kalman gain with an attention mechanism, enabling multimodal observation filtering with task-specific gain mechanisms. The proposed attention-based Multimodal Differentiable Filters ($\alpha$-MDF)~\cite{liu2023alphamdf} are depicted in Fig.~\ref{fig:overview}, with each module being learnable and operating in latent space. We demonstrate that $\alpha$-MDF excels in learning high-dimensional representations of system dynamics and capturing intricate nonlinear relationships. Furthermore, $\alpha$-MDF effectively models the nonlinear dynamics of soft robots, consistently outperforming differentiable filter baselines by up to 45\%.

%

\vspace{-0.09in}
\section{Multimodal Differentiable Filters}
\label{sec:method}
\vspace{-0.09in}
We introduce a novel approach called \textbf{a}ttention-based \textbf{M}ultimodal \textbf{D}ifferentiable \textbf{F}ilters ($\alpha$-\textbf{MDF}), which combines differentiable filters with insights from transformer models. 
This approach performs state estimation and fusion of multiple sensor modalities in a unified and differentiable manner. 
We utilize an ensemble method for Bayesian filtering wherein each ensemble member represents a compact robot state. Figure~\ref{fig:overview} shows the procedural steps of how this compact representation, known as the latent state, is obtained and get updated. The filtering process includes two essential steps, namely \emph{prediction} and \emph{update}, both of which are also implemented through neural networks. Most importantly, we replace the Kalman gain step with an attention mechanism, which is trained to weigh observations against predictions based on the current context. 
We will see that both steps can be naturally integrated into a single \textbf{attention gain} (AG) module.

Let ${\bf X}_{0:N}$ denote the latent states with dimension $d_x$ of $N$ steps in $t$ with number of $E$ ensemble members, ${\bf X}_{0:N} = [ {\bf x}^{1}_{0:N}, \dots, {\bf x}^{E}_{0:N}]$, where $E \in \mathbb{Z}^+$. 

\textbf{Prediction step}: In this step, the state transition model takes the previous states with the current action, and predicts the next state. To this end, we leverage the capabilities of transformer-style neural networks~\cite{vaswani2017attention}. In addition, we generate a probability distribution over the posterior by implementing the state transition model as a stochastic neural network. Therefore,
we can use the following prediction step to update each ensemble member, given a sequence of latent states ${\bf X}_{t-N:t-1}$: 
    \begin{equation}
    \begin{aligned}\label{eq:1}
          {\bf x}^{i}_{t|t-N:t-1} & \thicksim  f_{\pmb {\theta}} ({\bf x}^{i}_{t|t-N:t-1}|{\bf a}_{t}, {\bf x}^{i}_{t-N:t-1}), \  \forall i \in E.
    \end{aligned}
   \end{equation}
Where $f_{\pmb {\theta}}$ is a transformer-style neural network with multiple attention layers. In our framework,
the latent state and the action at $t$ are processed by positional and type embedding layers~\cite{vaswani2017attention} prior to being fed into $f_{\pmb {\theta}}$.
Matrix ${\bf X}_{t|t-N:t-1} \in \mathbb{R}^{d_x \times E}$ holds the updated ensemble members which are propagated one step forward in latent space.
For simplicity, we represent ${\bf X}_{t|t-N:t-1}$ as ${\bf X}_{t}$ to denote the predicted state. Further elaboration on positional embeddings, type embeddings, and filter initialization can be found in Appendix~\ref{app:MDF-1} for more comprehensive details.


\begin{wrapfigure}{r}{0.55\textwidth}
\vspace{-0.1in}
  \begin{center}
    \includegraphics[width=0.55\textwidth]{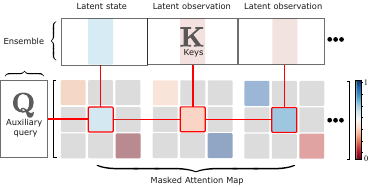}
  \end{center}
  \caption{Attention gain (AG) module uses a learned masked attention map to replace Kalman gain.}
  \label{fig:atten_gain}
  \vspace{-0.2in}
\end{wrapfigure}
\textbf{Update step}: A crucial step of the filtering process is the update step, which involves calculating the gain value. 
The proposed {\bf attention gain} (AG) module shown in Fig.~\ref{fig:atten_gain}, eliminates the need for an explicit observation model and can directly utilize high-dimensional features. By leveraging this approach, our framework enables a more flexible and efficient integration of measurements without the explicit requirement of a mapping function from the state to the measurement domain. Instead of using one sensor encoder, we use multiple sensor encoders $[s^{1}(\cdot), s^{2}(\cdot),\cdots,s^{M}(\cdot)]$ to learn latent observations from each modality:
\begin{align}\label{eq:sensor}
        \tilde{{\bf y}}^{(i, m)}_t  \thicksim  s^m (\tilde{{\bf y}}^{(i, m)}_t|{\bf y}^m_{t}),\  \forall \ i \in E, \ m \in M.
\end{align}
$M$ is the number of modalities in the system, $M \in \mathbb{Z}^+$. The encoders generate a series of latent observations, $\tilde{{\bf Y}}_t = [\tilde{{\bf Y}}^1_t, \cdots, \tilde{{\bf Y}}^M_t] \in \mathbb{R}^{M d_x \times E}$, where $\tilde{{\bf Y}}^m_t = [\tilde{{\bf y}}^{(1,m)}_t, \cdots, \tilde{{\bf y}}^{(E,m)}_t] \in \mathbb{R}^{d_x \times E}$. The latent observations are then concatenated with predicted state ${\bf X}_{t}$ as input to the AG model:
\begin{align}\label{eq:6}
     {\bf \hat{X}}_{t} =  \text{softmax}\left( \frac{{\pmb Q} ({\bf X}_{t}' \oplus \tilde{{\bf Y}}'_t )^T}{\sqrt{E}} \circ \pmb{\tilde{M}}  \right) ( {\bf X}_{t} \oplus \tilde{{\bf Y}}_t),
\end{align}
where ``$\oplus$'' denotes the concatenation and ``$\circ$'' is the Hadamard product, and ${\bf \hat{X}}_{t}$ is the final output. In general, an attention module typically receives three sequences of tokens: queries $\pmb Q$, keys $\pmb K$ and values $\pmb V$. In our case, we define $({\bf X}_{t}' \oplus \tilde{{\bf Y}}'_t)$ as the $\pmb K$ tokens, where ${\bf X}_{t}'$ and $\tilde{{\bf Y}}'_t$ are obtained by zero-centering, and the actual values of $({\bf X}_{t} \oplus \tilde{{\bf Y}}_t)$ are regarded as the $\pmb V$ tokens. 
As illustrated in Fig.~\ref{fig:atten_gain}, the length of the $\pmb K$ tokens is denoted as $d_k = (M+1) d_x$, where each token has a dimension of $E$, representing the distribution along this particular token index. 



\textbf{Placing Conditions on the Latent Space:} 
Within the framework of $\alpha$-MDF, we ensure consistency in the latent space by introducing a decoder model $\cal D$. This decoder model, implemented using multilayer perceptrons, projects the latent space onto the actual state space. By doing so, we resolve the alignment challenges in multimodal learning~\cite{liang2022foundations}, and gain meaningful comparisons when conducting sensor fusion and measurement update. Let ${\pmb x}_t$ be the ground truth state at $t$, the loss functions are defined as:
\begin{align}
\label{eq:loss1}
\mathcal{L}_{f_{\pmb {\theta}}} =  \| {\cal D}( f_{\pmb {\theta}}({\bf X}_{t}))- {\pmb x}_t\|_2^2,\ \ 
    \mathcal{L}_{\text{e2e}} =\| {\cal D}({\bf \hat{X}}_{t}) -  {\pmb x}_t\|_2^2,\ \ 
    \mathcal{L}_{s} = \|{\cal D} (s^m ({\bf y}^m_{t})) -{\pmb x}_t\|_2^2.
\end{align}
The final loss function is $\mathcal{L} = \mathcal{L}_{f_{\pmb {\theta}}} + \mathcal{L}_{\text{e2e}} + \mathcal{L}_{s}$, where $\mathcal{L}_{\text{e2e}}$ is the end-to-end loss. $\mathcal{L}_{f_{\pmb {\theta}}}$ is used to supervise the state transition model.  
The modular architecture of $\alpha$-MDF provides a key advantage in facilitating training and testing with masked modalities. The attention matrix $\pmb{\tilde{M}}$ can be disabled (set the attention values to zero) based on different input sensor modalities, thus improving the model's resilience to missing modalities.

\vspace{-0.09in}
\section{Experiments}\label{sec:result}
\vspace{-0.09in}

We conduct a series of experiments to evaluate the efficacy of the $\alpha$-MDF framework. 
Our study examines two categories of baselines: (a) DF baselines such as those proposed in~\cite{kloss2021train, jonschkowski2018differentiable, haarnoja2016backprop}, including dEKF~\cite{kloss2021train}, DPF~\cite{jonschkowski2018differentiable}, and dPF-M-lrn~\cite{kloss2021train}; and (b) sensor fusion baselines proposed in~\cite{lee2020multimodal}. Additional details on the baselines can be found in Appendix~\ref{app:MDF-3}. This experiment involves implementing the $\alpha$-MDF to model the dynamics of a soft robot system, especially Tensegrity robot~\cite{ikemoto2021development}.

\begin{figure*}[t!]
\centering
\includegraphics[width=\linewidth]{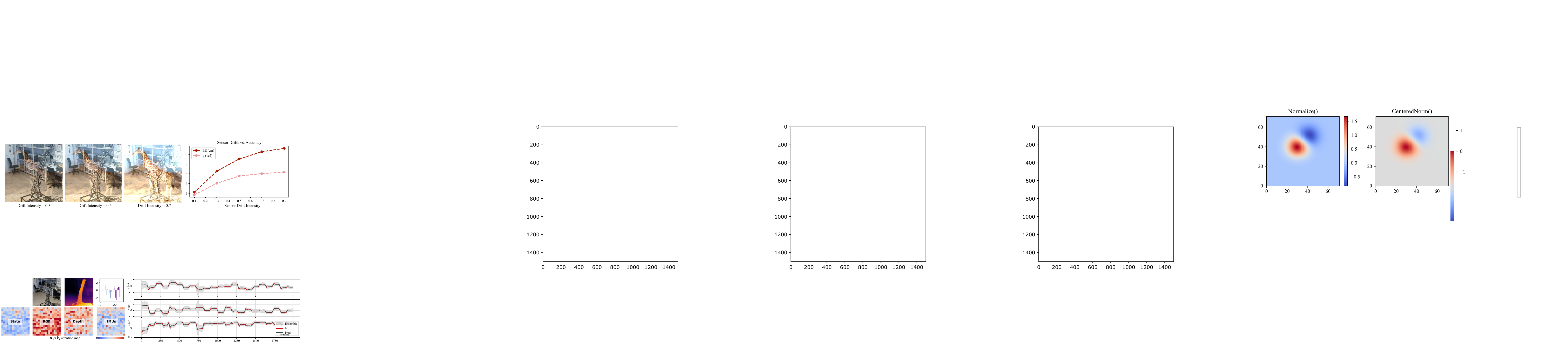}
\caption{\textbf{Left}: illustrate the learned attention gains corresponding to each modality and state. \textbf{Right}: the estimated positions of the end-effector (EE) for a tensegrity robot, represented as an ensemble distribution.}
\vspace{-1\baselineskip}
\label{fig:soft_result}
\end{figure*}

\textbf{Task Setup and Data:} The robot structure contains 5 layers of tensegrity. The actual state of a soft robot at time $t$ is represented by a 7-dimensional vector $\pmb {x}_t = [x, y, z, {\bf q}_x, {\bf q}_y, {\bf q}_z, {\bf q}_w]^T$, which denotes the position and orientation of the robot's hand tip.
The quaternion vector $\bf q$ represents the posture of the robot w.r.t the base (layer 1's bottom). In this task, we define ${\bf x} \in \mathbb{R}^{256}$ as the latent state. The complete set of modalities comprises [${\bf y}^1, {\bf y}^2, {\bf y}^3$], where ${\bf y}^1 \in \mathbb{R}^{224 \times 224 \times 3}$ represents RGB images, ${\bf y}^2 \in \mathbb{R}^{224 \times 224}$ is depth maps, and ${\bf y}^3 \in \mathbb{R}^{30}$ is proprioceptive inputs (IMUs). The action ${\bf a}_t$ of the system is the pressure vector of the 40 pneumatic cylinder actuators, where ${\bf a}_t \in \mathbb{R}^{40}$. In this experiment, synthetic depth maps are generated offline using the DPT model~\cite{ranftl2021vision}. Figure.~\ref{fig:soft_result} shows the recordings of RGB and the depth modalities, further details regarding the task setup and data collection is in Appendix~\ref{app:soft}.

\textbf{Results:} The soft robot modeling task is evaluated using 10-fold cross-validation and the mean absolute error (MAE) metric, and the results are presented in Table~\ref{Tab:soft_robot_result}. Our results demonstrate that $\alpha$-MDF outperforms the state-of-the-art methods in terms of DFs, achieving a MAE of 8.99cm. Specifically, our approach yields an MAE on the end-effector (EE) position estimation that is 45\%, 34\%, and 29\% lower than that obtained by dEKF, DPF, and dPF-M-lrn, respectively. 
\begin{wraptable}{r}{8cm}
\caption{Result evaluation on soft robot modeling task.}
\label{Tab:soft_robot_result}
\begin{center}
\scalebox{0.72}{
\begin{tabular}{c c c c c c}
    \toprule
     &RGB
     & Depth
     & IMUs
     & EE (cm)
     & {\bf q} ($10^1$)
     \\
    \midrule
    dEKF~\cite{kloss2021train}
     &
     & 
     & \checkmark
     &16.38$\pm$0.10
     &1.01$\pm$0.03
     \\
     DPF~\cite{jonschkowski2018differentiable}
     &
     & 
     & \checkmark
     &13.68$\pm$0.02
     &0.96$\pm$0.03
     \\
     dPF-M-lrn~\cite{kloss2021train}
     &
     & 
     & \checkmark
     &12.66$\pm$0.09
     &1.10$\pm$0.03
     \\
    \cellcolor{green!10}$\alpha$-MDF
     &\cellcolor{green!10}
     & \cellcolor{green!10}
     & \cellcolor{green!10}\checkmark
     &\cellcolor{green!10}\bf 8.99$\pm$0.02
     &\cellcolor{green!10}\bf 0.79$\pm$0.03
     \\
     \midrule
     Feature Fusion~\cite{lee2020multimodal}
     &\checkmark
     & \checkmark
     & \checkmark
     &8.35$\pm$0.22
     &0.60$\pm$0.03
     \\
     Unimodal~\cite{lee2020multimodal}
     &\checkmark
     & \checkmark
     & \checkmark
     &2.78$\pm$0.05
     &0.25$\pm$0.02
     \\
    Crossmodal~\cite{lee2020multimodal}
     &\checkmark
     & \checkmark
     & \checkmark
     &2.14$\pm$0.05
     &0.15$\pm$0.02
     \\
     \cellcolor{green!10}$\alpha$-MDF
     &\cellcolor{green!10}\checkmark
     & \cellcolor{green!10}\checkmark
     & \cellcolor{green!10}\checkmark
     &\cellcolor{green!10}\bf 1.67$\pm$0.09
     &\cellcolor{green!10}\bf 0.12$\pm$0.01
     \\
\bottomrule
\multicolumn{6}{l}{Means$\pm$standard errors.} \\
\end{tabular}}
\end{center}
\vspace{-0.2in}
\end{wraptable}
Of the sensor fusion baselines, crossmodal fusion~\cite{lee2020multimodal} exhibits marginally better outcomes than others, although it do not show any advantages over $\alpha$-MDF in predicting EE positions (2.14cm$\rightarrow$1.67cm). Notably, $\alpha$-MDF surpasses the feature fusion strategy by a significant margin of 4-fold. Additionally, appendix~\ref{app:soft} delves into an exploration of the potential benefits of modality selection for state estimation, where optimal combinations can be selected to achieve even higher accuracy. The results presented in Fig.~\ref{fig:soft_result} demonstrate the efficacy of $\alpha$-MDF in accurately estimating the state of soft robots in a multimodal setting, the ensemble distribution is indicated by gray shade representing the model uncertainty. With stable performance achieved over an extended duration of inference, $\alpha$-MDF has shown the potential in modeling dynamics for various complex non-linear systems.


\vspace{-0.1in}
\section{Conclusion}
\label{sec:conclusion}
\vspace{-0.09in}
This paper demonstrates the efficacy of using attention as a gain mechanism in differentiable Bayesian filtering and multimodal learning to improve the accuracy of learning soft robot dynamics. Experimental results highlight that $\alpha$-MDF enables soft robots to effectively handle high-dimensional observations and outperforms state-of-the-art baselines by up to 4-fold. Additionally, $\alpha$-MDF allows for investigating the importance of individual sensor modalities, showcasing its potential as a versatile tool for modeling soft robots.


\clearpage
\acknowledgments{This work was partially supported by JSPS KAKENHI Grant Numbers 22H03671 and 22K19815. We would like to sincerely acknowledge the valuable comments and feedback provided by the reviewers. Our gratitude also goes to Yuhei Yoshimitsu for assisting in the data collection with the tensegrity robot.}


\bibliography{example}  

\clearpage

\begin{appendices}

\section{Details in $\alpha$-MDF}\label{app:MDF}
This section provides a detailed overview of the previously mentioned $\alpha$-MDF modules, and describes differentiable Ensemble Kalman filters as the underlying DFs framework for $\alpha$-MDF.

\subsection{Model Initialization and Embedding Functions}\label{app:MDF-1}
An auxiliary model $\cal A$ is supplied in the filtering process to support training by starting the filter via projecting the actual state ${\pmb x}_{t-N:t-1}$ from low-dimensional space to latent space. The model is implemented using stochastic neural networks (SNNs)~\cite{jospin2022},
    \begin{equation}
    \begin{aligned}\label{eq:support}
          {\bf x}^{i}_{t-N:t-1} & \thicksim  {\cal A} ({\bf x}^{i}_{t-N:t-1}|{\pmb x}_{t-N:t-1}),\  \forall i \in E,
    \end{aligned}
   \end{equation}
where ${\bf x}^{i}_{t-N:t-1}$ is one latent state, the latent state ensemble is obtained by sampling $\cal A$ for $E$ times. During inference, we employ the trained sensor encoders' output, which is the latent representation of RGB, depth, or proprioception, as the initial state to initiate the filtering process.

Regarding the prediction step of $\alpha$-MDF, we apply positional embedding layers (sinusoidal functions)~\cite{vaswani2017attention} in the transformer process model (Eq.~\ref{eq:1}) to generate ${\pmb{e}}_{t-N:t-1}$ as the embedding for time-series data, ${\pmb{e}}_{t-N:t-1} = f_{\mathcal{L}}({\bf X}_{t-N:t-1}) \in \mathbb{R}^{d_x \times (N-1)}$. The positional embedding layer is utilized to label the state by index it with time $t$. When activating the action ${\bf a}_t$ in the process model, we also utilize a type embedding layer that indexes ${\pmb{e}}_{t-N:t-1}$ and ${\bf a}_t$ with 0 and 1, and then fed to sinusoidal functions. Subsequently, the element-wise summation of outputs obtained from the aforementioned procedures serve as input to the transformer process model for further processing.

\subsection{Differentiable Ensemble Kalman Filter}\label{app:MDF-2}
Unlike prior proposals for differentiable filters, such as dEKF~\cite{kloss2021train} and DPF~\cite{jonschkowski2018differentiable}, Differentiable Ensemble Kalman Filter~\cite{evensen2003ensemble} leverages recent advancements in stochastic neural networks (SNNs)~\cite{jospin2022}. Specifically, we draw inspiration from the work in~\cite{gal2016dropout}, which established a theoretical connection between the Dropout training algorithm and Bayesian inference in deep Gaussian processes. As a result, we can use stochastic forward passes to produce empirical samples from the predictive posterior of a neural network trained with Dropout.
Hence, for the purposes of filtering, we can implicitly model the process noise by sampling state from a neural network trained on the transition dynamics, i.e., ${\bf x}_{t}  \thicksim  f_{\pmb {\theta}} ({\bf x}_{t-1})$. In contrast to previous approaches~\cite{jonschkowski2018differentiable, kloss2021train}, the transition network $f_{\pmb {\theta}}(\cdot)$ models the system dynamics, as well as the inherent noise model in a consistent fashion without imposing diagonality.

\textbf{Prediction Step}: Similar to $\alpha$-MDF, we use an initial ensemble of $E$ members to represent the initial state distribution ${\bf X}_0 = [ {\bf x}^{1}_0, \dots, {\bf x}^{E}_0]$, $E \in \mathbb{Z}^+$. We leverage the stochastic forward passes from a trained state transition model to update each ensemble member: 
    \begin{equation}
    \begin{aligned}\label{eq:add1}
          {\bf x}^{i}_{t|t-1} & \thicksim  f_{\pmb {\theta}} ({\bf x}^{i}_{t|t-1}|{\bf x}^{i}_{t-1|t-1}),\  \forall i \in E.
    \end{aligned}
   \end{equation}
 Matrix ${\bf X}_{t|t-1} = [{\bf x}^{1}_{t|t-1}, \cdots, {\bf x}^{E}_{t|t-1}]$ holds the updated ensemble members which are propagated one step forward through the state space. Note that sampling from the transition model $f_{\pmb {\theta}}(\cdot)$ (using the SNN methodology described above) implicitly introduces a process noise.
 
\textbf{Update step}: 
Given the updated ensemble members ${\bf X}_{t|t-1}$, a nonlinear observation model $h_{\pmb {\psi}}(\cdot)$ is applied to transform the ensemble members from the state space to observation space. Following our main rationale, the observation model is realized via a neural network with weights $\pmb {\psi}$. Accordingly, the update equations become:

\noindent\begin{minipage}{.65\linewidth}
\begin{equation}\label{eq:add2}
        {\bf H}_t {\bf A}_{t} = {\bf H}_t {\bf X}_{t}
        - \left[\frac{1}{E} \sum_{i=1}^E h_{\pmb {\psi}}({\bf x}^i_{t}),
        \cdots,
        \frac{1}{E} \sum_{i=1}^E h_{\pmb {\psi}}({\bf x}^i_{t})\right],
\end{equation}
\end{minipage}
\begin{minipage}{.35\linewidth}
\begin{equation}\label{eq:add3}
        \tilde{{\bf y}}^{i}_t  \thicksim  s (\tilde{{\bf y}}^{i}_t|{\bf y}_{t}),\  \forall \ i \in E.
\end{equation}
\end{minipage}
   
${\bf H}_t {\bf X}_{t}$ is the predicted observation, and ${\bf H}_t {\bf A}_{t}$ is the sample mean of the predicted observation at $t$.
Traditional Ensemble Kalman Filter treats observations as random variables. Hence, the ensemble can incorporate a measurement perturbed by a small stochastic noise to reflect the error covariance of the best state estimate~\cite{evensen2003ensemble}.
In differentiable Ensemble Kalman Filter, we incorporate a Bayesian sensor encoder $s(\cdot)$.
Sensor encoder serves to learn projections from observation space to latent space as in Eq.~\ref{eq:add3},
where ${\bf y}_{t}$ represents the noisy sensor observation. Sampling from sensor encoder yields latent observations
$\tilde{{\bf Y}}_t = [\tilde{{\bf y}}^{1}_t, \cdots, \tilde{{\bf y}}^{E)}_t]$.
The KF update step can then be continued by using the learned observation and predicted observation:
\begin{equation}
\begin{aligned}\label{eq:add4}
    {\bf K}_t = \frac{1}{E-1} {\bf A}_t ({\bf H}_t {\bf A}_t)^T (\frac{1}{E-1}  ({\bf H}_t {\bf A}_t)  ({\bf H}_t {\bf A}_t)^T + {\bf R})^{-1}.
\end{aligned}
\end{equation}
The measurement noise model ${\bf R}$ is implemented using a multilayer perceptron (MLP), similar to the implementation in~\cite{kloss2021train}. The MLP takes a learned observation $\tilde{{\bf Y}}_t$ at time $t$ and produces a noise covariance matrix. The final estimate of the ensemble ${\bf \hat{X}}_{t}$ is obtained by performing the measurement update step, given by:
\begin{align}\label{eq:add5}
    {\bf \hat{X}}_{t} &= {\bf X}_{t} + {\bf K}_t (\tilde{{\bf Y}}_t - {\bf H}_t {\bf X}_{t}).
\end{align}
In inference, the ensemble mean ${\bf \bar{x}}_{t|t} = \frac{1}{E}\sum_{i=1}^E {\bf x}^i_{t|t}$ is used as the updated state.

\subsection{Baselines}\label{app:MDF-3}
In our study, we examine two categories of baselines: (a) DFs baselines, which consist of existing methods such as those proposed in~\cite{kloss2021train, jonschkowski2018differentiable, haarnoja2016backprop}, and (b) sensor fusion strategies, as proposed in~\cite{lee2020multimodal}.

\begin{table}[h]
\caption{Dimensions pertinent to robot state estimation task.}
\label{Tab:baseline_dim}
\begin{center}
\scalebox{0.82}{
\begin{tabular}{c c c c}
    \toprule
     \multirow{2}{3em}{Method} 
     &\multicolumn{3}{c}{{\bf Soft Robot}}
     \\
     &State
     &Observation
     &Action
     \\
    \midrule
     dEKF~\cite{kloss2021train}
     &7
     &7
     &40
     \\
    DPF~\cite{jonschkowski2018differentiable}
     &7
     &7
    &40
     \\
    dPF-M-lrn~\cite{kloss2021train}
     &7
     &7
     &40
     \\
     \midrule
     Feature Fusion~\cite{lee2020multimodal}
     &7
     &7
     &40
     \\
    Unimodal~\cite{lee2020multimodal}
     &7
     &7
     &40
     \\
     Crossmodal~\cite{lee2020multimodal}
     &7
     &7
     &40
     \\
    \cellcolor{green!10} $\alpha$-MDF
     &\cellcolor{green!10}256
     &\cellcolor{green!10}256
     &\cellcolor{green!10}40
     \\
\bottomrule
\end{tabular}}
\end{center}
\vspace{-0.1in}
\end{table}

\textbf{Dimensionality:} Table~\ref{Tab:baseline_dim} presents the dimensions for the state, observations, and actions utilized for each of the tasks. To ensure consistency, we opt for a dimension of 256 for $\alpha$-MDF in all tasks, thus, enabling filtering over high-dimensional spaces. Unlike the baseline methods, which use low-dimensional state definitions, we filter over higher dimension spaces with $\alpha$-MDF.

\textbf{Differentiable Filters:} To maintain consistency in the comparison of results against the DFs baselines, we train $\alpha$-MDF with a single modality. The baselines in this category include the differentiable Extended Kalman filter (dEKF)~\cite{kloss2021train}, differentiable particle filter (DPF)~\cite{jonschkowski2018differentiable}, and the modified differentiable particle filter (dPF-M-lrn)~\cite{kloss2021train}, which uses learned process and process noise models. For dEKF, the Jacobian matrix in the prediction step can either be learned end-to-end or supplied if the motion model is known. DPF employs 100 particles for both training and testing and also incorporates an observation likelihood estimation model $l$. This module takes in an image embedding and produces a likelihood that updates each particle's weight. Unlike DPF, dPF-M-lrn implements a learnable process noise model. It also adopts a Gaussian Mixture Model for calculating the likelihood for all particles. It is worth noting that all the baseline methods perform Kalman filtering on low-dimensional actual state space, whereas $\alpha$-MDF executes the filtering process in the latent space.

\textbf{Sensor Fusion:} Regarding sensor fusion baselines, we use three strategies discussed in~\cite{lee2020multimodal}, namely, Feature Fusion, Unimodal Fusion, and Crossmodal Fusion. The Feature Fusion strategy aims to process each modality individually and subsequently merge the modalities to generate a multimodal feature set using neural networks, which is then used for state estimation. The Unimodal Fusion treats each modality $\mathcal{N}\sim(\pmb{\mu}_t^{M_1}, \pmb{\Sigma}_t^{M_1})$ and $\mathcal{N}\sim(\pmb{\mu}_t^{M_2}, \pmb{\Sigma}_t^{M_2})$ as distributions and fuse two unimodal distribution as one normally distributed multimodal distribution $\mathcal{N}\sim(\pmb{\mu}_t, \pmb{\Sigma}_t)$:
\begin{align}\label{eq:unimodal}
\pmb{\mu}_t = \frac{(\pmb{\Sigma}_t^{M_1})^{-1} \pmb{\mu}_t^{M_1} + (\pmb{\Sigma}_t^{M_2})^{-1} \pmb{\mu}_t^{M_2}}{(\pmb{\Sigma}_t^{M_1})^{-1}+(\pmb{\Sigma}_t^{M_2})^{-1} }, \ \ 
    \pmb{\Sigma}_t = ((\pmb{\Sigma}_t^{M_1})^{-1}+(\pmb{\Sigma}_t^{M_2})^{-1})^{-1},
\end{align}
where the associative property can be used for fusing more than two modalities. For Crossmodal Fusion, information from one modality can be used to determine the uncertainty of the other ones, two coefficients are proposed as $\pmb{\beta}_t^{M_1}$ and $\pmb{\beta}_t^{M_2}$, where each coefficient has the same dimension of the state, the fused distribution is:
\begin{align}\label{eq:crossmodal}
\pmb{\mu}_t = \frac{\pmb{\beta}_t^{M_1} \circ \pmb{\mu}_t^{M_1} + \pmb{\beta}_t^{M_2} \circ \pmb{\mu}_t^{M_2}}{\pmb{\beta}_t^{M_1} + \pmb{\beta}_t^{M_2}}, \ \ 
    \pmb{\Sigma}_t = \frac{\pmb{B}_t^{M_1} \circ \pmb{\Sigma}_t^{M_1} + \pmb{B}_t^{M_2} \circ \pmb{\Sigma}_t^{M_2}} {\pmb{B}_t^{M_1} + \pmb{B}_t^{M_2}},
\end{align}
where $\pmb{B}_t^{M} = (\pmb{\beta}_t^{M})^T \pmb{\beta}_t^{M}$. As mentioned in~\cite{lee2020multimodal}, each sensor encoder was independently trained and subsequently used for end-to-end training with DFs. We adopt a similar approach, but with a differentiable Ensemble Kalman Filter backbone in place instead. The resampling procedure from the fused distribution in this scenario is achieved by using the reparematerization trick~\cite{kingma2015variational}.

\subsubsection{Regarding Attention Gain}\label{MDF-4} 
In a traditional attention mechanism, the proximity of $\pmb Q$ and $\pmb K$ is measured, and $\pmb V$ that is associated with $\pmb K$ is utilized to generate outputs. However, we posit that within each latent vector, every index is probabilistically independent, and index $i$ of a latent state should only consider index $i$ of each latent observation. To accomplish this, we utilize matrix $\pmb{\tilde{M}}$ to retain only the diagonal elements of each $(d_x \times d_x)$ attention map, which enforces the attention weights to be determined according to the corresponding indices. As depicted in Fig.~\ref{fig:atten_gain}, the red line represents the mapping for a single latent state token index.
Auxiliary query tokens $\pmb Q \in \mathbb{R}^{d_x \times E}$ are introduced as trainable parameters in the neural network to facilitate learning. It is important to note that both the $\pmb Q$ and $\pmb K$ tokens undergo positional embedding before being fed into the AG module.

To empirically verify this assumption, we conducted an additional experiment where we trained an alternative $\alpha$-MDF framework. In this framework, we deliberately excluded the matrix $\pmb{\tilde{M}}$ in the attention gain module for the specific purpose of evaluating the effect soft robot modeling task. The results show a significant increase in the Mean Absolute Error (MAE) for end-effector estimation when the masked map $\pmb{\tilde{M}}$ was excluded from the attention gain module. Specifically, the MAE of the end-effector increased from 1.67cm to 5.23cm. Based on these results, it is strongly recommended to utilize the masked map $\pmb{\tilde{M}}$ within the attention gain module for improved performance in both joint angle estimation and end effector tracking.


\section{Additional Experiments}
This section presents supplementary experimental results. For Soft Robot Modeling Tasks, we concentrate mainly on ablation studies.


\subsection{Soft Robot Modeling Tasks}\label{app:soft}
\begin{figure*}[h]
\centering
\includegraphics[width=0.7\linewidth]{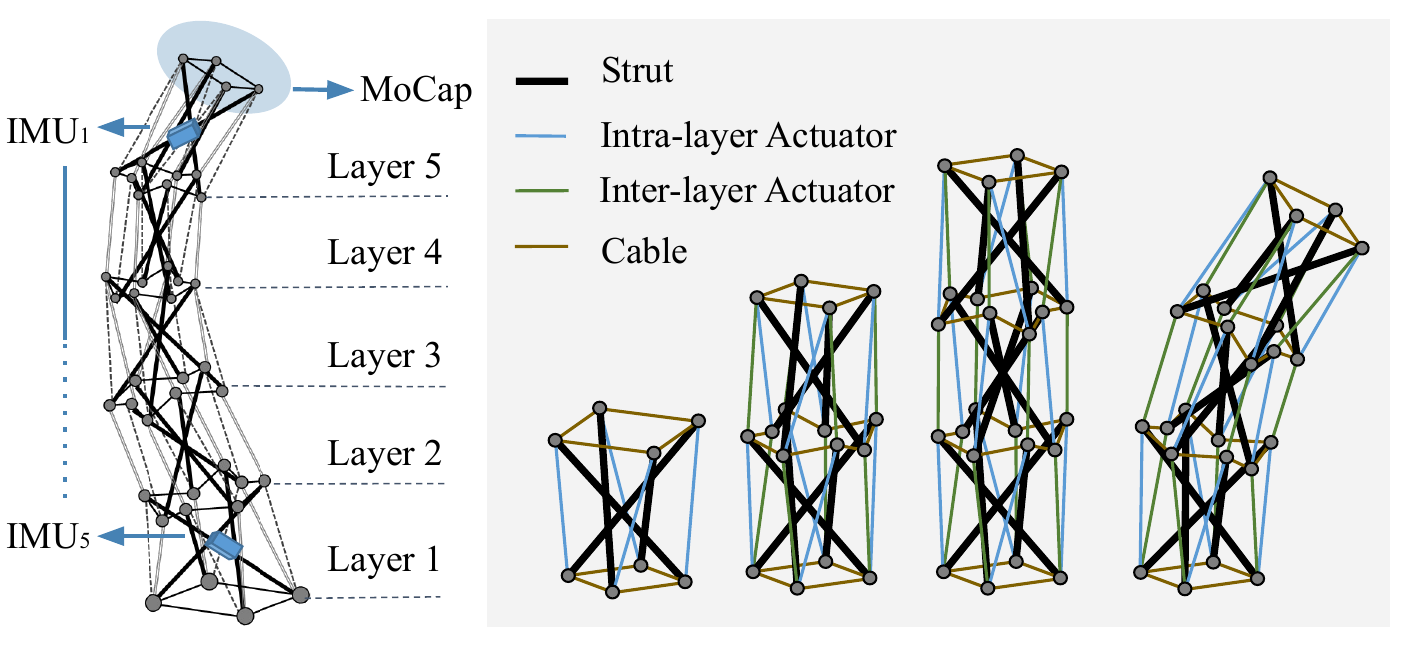}
\caption{The tensegrity robot features 5 flexible layers, each a tensegrity module with struts, cables, and actuators.}
\vspace{-1\baselineskip}
\label{fig:structure}
\end{figure*}

This section presents a comprehensive analysis of the tensegrity robot structure, the bending motion mechanism, and pertinent sensory information, followed by a description of additional experimental outcomes related to this task.

\textbf{Preliminaries}:
Our research utilizes a tensegrity robot arm (developed in~\cite{ikemoto2021development}) that follows a strict tensegrity structure featuring struts, cables (including spring-loaded and actuated cables), and five layers of arm-like tensegrity structures, which produce continuous bending postures when exposed to external forces. The longitudinal length is maintained by stiff cables, while the bending direction is solely determined by external forces. We determine the robot's kinematics through data from Inertial Measurement Units (IMUs), optical motion capture (MoCap), and proportional pressure control valves, with each of the five struts in each layer featuring an IMU. We also record the video by placing a camera in front of the robot while collecting all sensory data.

A soft robot's state at $t$ is a 7-dimensional vector ${\bf x}_t = [x, y, z, {\bf q}_x, {\bf q}_y, {\bf q}_z, {\bf q}_w]^T$, indicating its position and orientation relative to the base frame (layer 1's bottom). ${\bf q}$ represents the robot's posture. The system's action is the pressure vector of its 40 pneumatic cylinder actuators (${\bf a}_t \in \mathbb{R}^{40}$). Its raw observation is comprised of 5 IMU readings (${\bf y}^3_t \in \mathbb{R}^{30}$), with each IMU measuring a 6-dimensional vector of accelerations and angular velocities relative to its location. Fig.~\ref{fig:structure} illustrates the locations of the IMUs on the struts (blue cubes) in each layer.

\begin{figure*}[t]
\centering
\includegraphics[width=\linewidth]{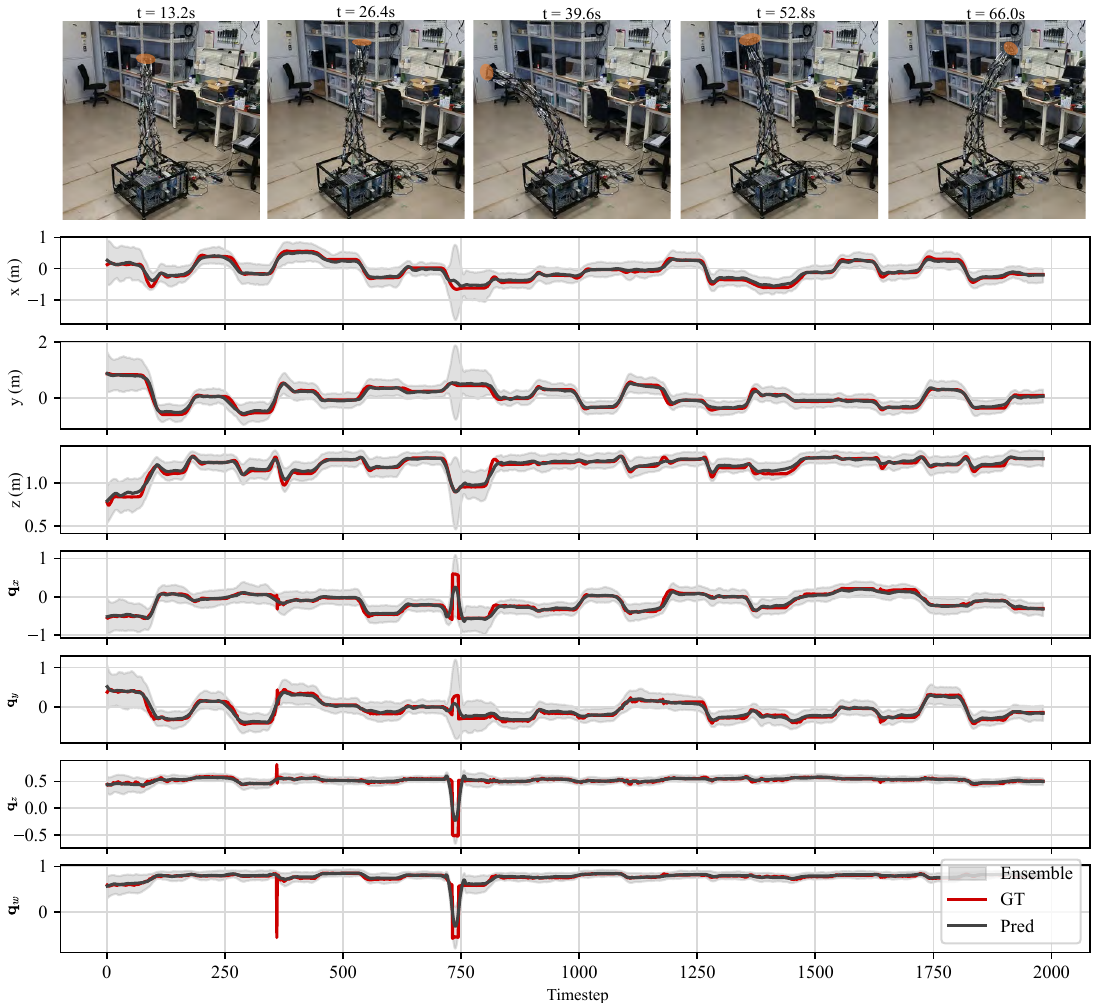}
\caption{Predicted end-effector (EE) positions and quaternion vectors $\bf q$ in the soft robot modeling task. The \textbf{top} row displays the actual robot posture at the corresponding time, with the orange circle indicating the EE positions, which are not included in the RGB modality input.}
\vspace{-1\baselineskip}
\label{fig:soft_full}
\end{figure*}

\textbf{Data}: The complete set of modalities comprises [${\bf y}^1, {\bf y}^2, {\bf y}^3$], where ${\bf y}^1 \in \mathbb{R}^{224 \times 224 \times 3}$ represents RGB images, ${\bf y}^2 \in \mathbb{R}^{224 \times 224}$ is synthetic depth maps which we generate from DPT repo~\cite{ranftl2021vision} utilizing ``Intel/dpt-large'', and ${\bf y}^3 \in \mathbb{R}^{30}$ is proprioceptive inputs (IMUs).
The dataset is generated by performing optical motion capture on the real tensegrity robot hand tip while randomly supplying desired pressure vectors to the pneumatic cylinder actuators. The action ${\bf a}_t \in \mathbb{R}^{40}$, 
5 IMU readings ${\bf y}^3_t \in \mathbb{R}^{30}$, 
and a 7-dimensional state ${\bf x}_t$ are recorded,
with 40-dimensional pressure vectors being used as a control signal. 
A total of 12,000 trials of robot motion are collected, with each trial involving moving the robot from its current equilibrium posture to the next equilibrium posture by applying the new desired pressure. All data are collected via a ROS2 network with a sampling frequency of 30Hz and are synchronized using the ``message\_filters'' package.

\begin{wraptable}{r}{8cm}
\caption{Ablation study on Tensegrity robot.}
\label{Tab:soft_robot_ablation}
\begin{center}
\scalebox{0.87}{
\begin{tabular}{c c c c c}
    \toprule
     RGB
     & Depth
     & IMUs
     & EE (cm)
     & {\bf q}($10^1$)
     \\
    \midrule
     \checkmark
     & 
     & 
     &2.07$\pm$0.03
     &0.31$\pm$0.08
     \\
     & \checkmark
     & 
     &2.77$\pm$0.01
     &0.19$\pm$0.05
     \\
     \cellcolor{red!10}
     & \cellcolor{red!10}
     & \cellcolor{red!10}\checkmark
     &\cellcolor{red!10}8.99$\pm$0.02
     &\cellcolor{red!10}0.79$\pm$0.03
     \\
     & \checkmark
     & \checkmark
     &2.08$\pm$0.03
     &0.14$\pm$0.02
     \\
     \checkmark
     & 
     & \checkmark
     & 1.73$\pm$0.05
     & 0.12$\pm$0.02
     \\
     \checkmark
     & \checkmark
     & 
     &1.74$\pm$0.06
     &\bf 0.10$\pm$0.02
     \\
     \cellcolor{green!10}\checkmark
     & \cellcolor{green!10}\checkmark
     & \cellcolor{green!10}\checkmark
     &\cellcolor{green!10}\bf 1.67$\pm$0.09
     &\cellcolor{green!10}0.12$\pm$0.01
     \\
\bottomrule
\multicolumn{5}{l}{Means$\pm$standard errors.} \\
\end{tabular}}
\end{center}
\vspace{-0.2in}
\end{wraptable}

\subsection{Ablation Study} 
\begin{figure*}[h]
\centering
\includegraphics[width=\linewidth]{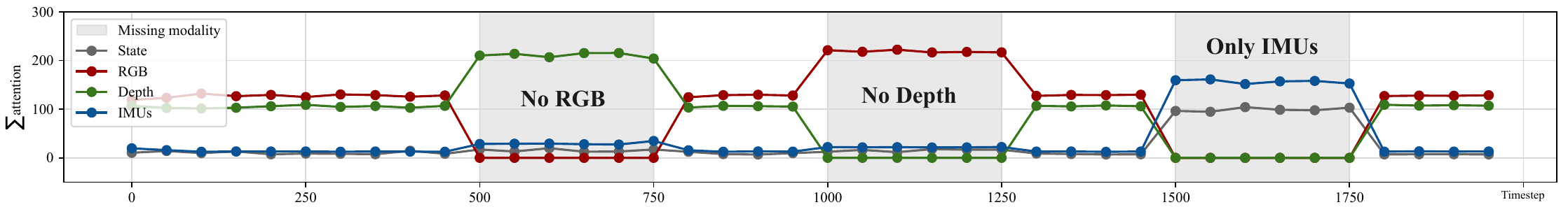}
\caption{The corresponding accumulated attention values for each modality during testing. The gray areas show certain modalities are selected or not selected.}
\vspace{-1\baselineskip}
\label{fig:soft_atten}
\end{figure*}

In addition to the results presented in Section~\ref{sec:result}, we evaluate various combinations of modalities to determine whether an optimal subset of modalities can be identified to attain comparable outcomes without using all modalities during the filtering operation. As demonstrated in Table\ref{Tab:soft_robot_ablation}, utilizing only one modality fails to achieve comparable results, with the highest accuracy (2.07cm) exclusively from employing ${\bf y}^1$ (RGB). The lowest error in posture estimation for the robot is obtained by leveraging [${\bf y}^1, {\bf y}^2$], showing slight improvement (0.10$\rightarrow$0.12) over leveraging the full modalities [${\bf y}^1, {\bf y}^2, {\bf y}^3$]. However, the lowest MAE error for the EE position persists even when all modalities are employed. Interestingly, using solely ${\bf y}^3$ results in the highest state estimation error, which aligns with the lowest attention value visualized in Fig~\ref{fig:soft_atten}. As depicted in Fig.~\ref{fig:soft_atten}, it is evident that $\alpha$-MDF prioritizes ${\bf y}^1$ over other modalities. Interestingly, the attention values change upon turning off certain modalities while the system remains stable and functional.

\subsection{Concept Drifts}
\begin{figure*}[h]
\centering
\includegraphics[width=\linewidth]{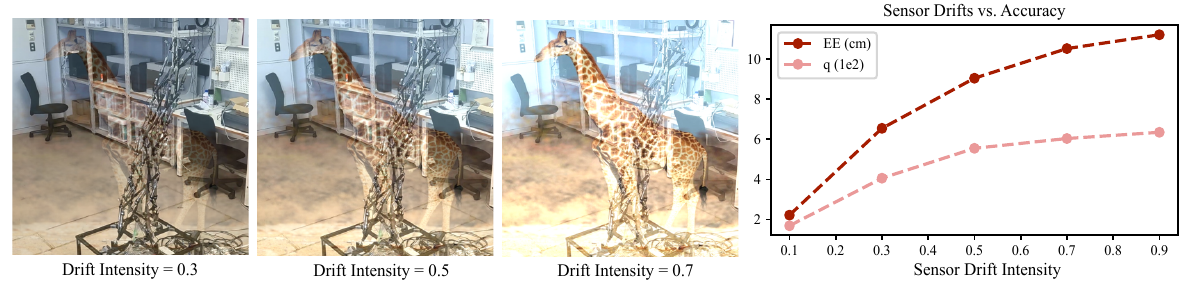}
\caption{Concept drifts analysis by adding background change with scale in RGB space.}
\vspace{-1\baselineskip}
\label{fig:concept_drifts}
\end{figure*}

To investigate the effects of concept drift and contextual changes~\cite{lu2018learning} on the $\alpha$-MDF framework, we incorporated a background change at inference time. In particular, image blending is used to overlay a different RGB picture into the background. The objective of the experiment is to inference behavior when changes to the environment occur. We evaluated the tracking performance at various blending levels, as illustrated in Fig.~\ref{fig:concept_drifts}. The results provide an understanding of how effectively the $\alpha$-MDF framework handles concept drift at different levels of intensity. It is noteworthy that despite substantially affecting the visual representation of the scene the achieved results (6.54cm) are comparable to utilizing only IMUs (8.99cm). This experiment provides an early insight into the utility of multimodality for mitigating the adverse effects due to contextual changes and concept drift.

\section{Complexity and Training Details}
In this section, we present an analysis of the computational complexity associated with each task by measuring the wall-clock time. Additionally, we provide comprehensive information regarding the model hyper-parameters and training curriculum employed for the experiments. These details offer insights into the computational requirements and settings utilized for training the models in our study.

\subsection{Complexity}
To assess the computational complexity of the proposed $\alpha$-MDF framework alongside the baseline differentiable filters (DFs), we measured the wall-clock time during inference. The results, provided in Table~\ref{Tab:complexity}, demonstrate the computational time for each approach. In the comparison with DFs baselines, we only considered a single modality. It is worth noting that in the multimodality setting, we observed only a marginal increase in the elapsed time (0.03 sec) when handling multiple types of observations. This indicates that the proposed framework, $\alpha$-MDF, is efficient and capable of effectively handling various modalities without significantly compromising computational performance.

\begin{table*}[h]
\caption{Wall-clock time (sec) for each task.}
\label{Tab:complexity}
\begin{center}
\scalebox{0.82}{
\begin{tabular}{c c c }
    \toprule
    &Modality
     &Soft Robot Task
     \\
    \midrule
    dEKF~\cite{kloss2021train}
    & 1
    &0.0474$\pm$0.003
     \\
    DPF~\cite{jonschkowski2018differentiable}
    & 1
    &0.0600$\pm$0.004
     \\
    dPF-M-lrn~\cite{kloss2021train}
    & 1
    &0.0590$\pm$0.002
     \\
    $\alpha$-MDF
    & 1
    &0.0633$\pm$0.003
     \\
    $\alpha$-MDF
    & $\geq$2
    &0.0910$\pm$0.004
     \\
\bottomrule
\multicolumn{3}{l}{Means$\pm$standard errors.} \\
\end{tabular}}
\end{center}
\vspace{-0.2in}
\end{table*}

\subsection{Training Details}

Table~\ref{tab:MDF_module} provides an exhaustive enumeration of all learnable modules utilized in $\alpha$-MDF, which includes three primary components: the state transition model $f_{\pmb {\theta}}$, the sensor encoders $[s^{1}(\cdot), s^{2}(\cdot),\cdots,s^{M}(\cdot)]$, and the attention gain (AG) module. We adopt self-attention layers with dimension 256 and 8 heads, denoted as ``Self Attn'', in the state transition model. The cross-attention layers, denoted as ``Cross Attn'', is with dimension 32 and 4 heads in the AG module. The sensor encoders utilized in our approach and all baseline models are identical, with $s^1$ acting on image-like modalities, utilizing ResNet18~\cite{he2016identity} for learning high-dimensional observation representations, while $s^2$ pertains to low-dimensional modalities such as joint angles. The auxiliary model $\mathcal{A}$ and the decoder $\mathcal{D}$ shares a similar structure to $s^2$, but with different number of neurons. 
Note that $x$ is the dimension of the actual state.

\begin{table}[h]
  \centering
  \vspace{-0.1in}
  \caption{$\alpha$-MDF's learnable sub-modules.}
  \label{tab:MDF_module}
  \scalebox{0.84}{
  \begin{tabular}{ll}
    \toprule
\vspace{+0.1in}
$f_{\pmb {\theta}}$: & 3$\times$ SNN(256, ReLu), Positional Embedding, 3$\times$ Self Attn(256,8), 2$\times$ SNN(256, ReLu), 1$\times$ SNN($d_x$, -) \\
\vspace{+0.1in}
$s^{1}$: & 1$\times$ ResNet18(h,w,ch), 2$\times$ fc(2048, ReLu), 1$\times$ SNN(512, ReLu), 1$\times$ SNN($d_x$, -)\\
\vspace{+0.1in}
$s^{2}$: & 1$\times$ SNN(128, ReLu), 1$\times$ SNN(256, ReLu), 1$\times$ SNN(512, ReLu), 1$\times$ SNN($d_x$, -)\\
\vspace{+0.1in}
AG: & Positional Embedding, 1$\times$ Cross Attn(32, 4, mask)\\
\vspace{+0.1in}
$\mathcal{A}$: & 1$\times$ SNN(128, ReLu), 1$\times$ SNN(256, ReLu), 1$\times$ SNN(512, ReLu), 1$\times$ SNN(1024, ReLu), 1$\times$ SNN($d_x$, -)\\

$\mathcal{D}$: & 1$\times$ fc(256, ReLu), 1$\times$ SNN(128, ReLu), 1$\times$ SNN(32, ReLu), 1$\times$ SNN($x$, -)\\
    \bottomrule
\multicolumn{2}{l}{fc: Fully Connected, SNN: Stochastic Neural network.} \\
  \end{tabular}}
  \vspace{-0.1in}
\end{table}

During $\alpha$-MDF training, we employ the curriculum outlined in Algorithm~\ref{alg:latent}. Note that some tasks may require pre-training the sensor encoders before performing end-to-end training the entire framework. For each task, we train $\alpha$-MDF model with utilizing batch size of 64 on a single NVIDIA A100 GPU for roughly 48 hours. For all the tasks, we use the Adamw~\cite{loshchilov2017decoupled} optimizer with a learning rate of 1e-4.

  \begin{algorithm}[H]
    \caption{Condition in Latent Space: training algorithm\label{alg:latent} return the weights $\omega$}
    \begin{algorithmic}
\State \textbf{Input:} \text{$\alpha$-MDF}, dataloader $\left( 
\{{\pmb x}_t\}_{t-N}^{t+1}, \{{\bf y}^m_{t}\}_{m=1}^M, \{{\bf y}^m_{t+1}\}_{m=1}^M, \{{\bf a}_{t}\}_{t-1}^{t+1}\right)$ 
\State \textbf{Output:} weights $\omega$
\While{not converged} 
\State $\text{ Call dataloader with a random timestep $t$.}$
\For{timestep $t \gets t$ to $t+1$}
\State $e_1 \gets \sum^{M}_{m=1} \|{\cal D} (s^m ({\bf y}^m_{t})) -{\pmb x}_t\|_2^2$ according to Eq.~\ref{eq:loss1}
\State $e_2 \gets  \mathcal{L}_{f_{\pmb {\theta}}}({\bf X}_{t}) + \mathcal{L}_{e2e}({\bf \hat{X}}_{t})$ according to Eq.~\ref{eq:loss1}
\State $e_t  \gets e_1 + e_2$ 
\EndFor 
\State $\omega  \gets \text{Train}\left(\text{$\alpha$-MDF}, e_t + e_{t+1}\right)$ 
\EndWhile
\State \textbf{return} $\omega$
    \end{algorithmic}
  \end{algorithm}

\end{appendices}

\end{document}